\documentclass[nohyperref]{article}

\usepackage{microtype}
\usepackage{graphicx}
\usepackage{subfigure}
\usepackage{booktabs} 

\usepackage{hyperref}

\usepackage[accepted]{icml2022}

\usepackage{amsmath}
\usepackage{amssymb}
\usepackage{mathtools}
\usepackage{amsthm}
\usepackage{soul}
\newcommand{\dashrule}[1][black]{%
  \color{#1}\rule[\dimexpr.5ex-.2pt]{4pt}{.4pt}\xleaders\hbox{\rule{4pt}{0pt}\rule[\dimexpr.5ex-.2pt]{4pt}{.4pt}}\hfill\kern0pt%
}

\usepackage[capitalize,noabbrev]{cleveref}

\theoremstyle{plain}

\theoremstyle{definition}

\theoremstyle{remark}

\allowdisplaybreaks

\def\onedot{.}
 
\def\ie{\emph{i.e}\onedot}

\usepackage[textsize=tiny]{todonotes}

\icmltitlerunning{IKNet}

\begin{document}

\twocolumn[
\icmltitle{Neural Inverse Kinematics }




\begin{icmlauthorlist}
\icmlauthor{Raphael Bensadoun}{comp}
\icmlauthor{Shir Gur}{comp}
\icmlauthor{Nitsan Blau}{comp}
\icmlauthor{Tom Shenkar}{comp}
\icmlauthor{Lior Wolf}{comp}
\end{icmlauthorlist}

\icmlaffiliation{comp}{Mentee Robotics}

\icmlcorrespondingauthor{Raphael Bensadoun}{raphael@menteebot.com}
\icmlcorrespondingauthor{Shir Gur}{shir@menteebot.com}

\icmlkeywords{Machine Learning, ICML}

\vskip 0.3in
]



\printAffiliationsAndNotice{} 

\begin{abstract}
Inverse kinematic (IK) methods recover the parameters of the joints, given the desired position of selected elements in the kinematic chain. While the problem is well-defined and low-dimensional, it has to be solved rapidly, accounting for multiple possible solutions. In this work, we propose a neural IK method that employs the hierarchical structure of the problem to sequentially sample valid joint angles conditioned on the desired position and on the preceding joints along the chain. In our solution, a hypernetwork $f$ recovers the parameters of multiple primary networks {$g_1,g_2,\dots,g_N$, where $N$ is the number of joints}, such that each $g_i$ outputs a distribution of possible joint angles, and is conditioned on the sampled values obtained from the previous primary networks $g_j, j<i$. The hypernetwork can be trained on readily available pairs of matching joint angles and positions, without observing multiple solutions. At test time, a high-variance joint distribution is presented, by sampling sequentially from the primary networks. We demonstrate the advantage of the proposed method both in comparison to other IK methods for isolated instances of IK and with regard to following the path of the end effector in Cartesian space.
\end{abstract}

\section{Introduction}
\label{sec:intro}

Given the joint angles, the position and orientation of the robot's end-effector can be readily computed in a process called forward-kinematics. However, robotic planning and controls require mapping in the other direction, \ie, from the end-effector's Cartesian space coordinates to the joint positions. This inverse mapping is called Inverse Kinematic (IK). It is a nonlinear problem that often has multiple solutions~\cite{craig2009introduction}.

For simple kinematic chains without much ambiguity, one can obtain analytical solutions for the IK problem. However, for the chain type that appears in robotic arms and other complex robots, one has to rely on numerical methods. In this work, we propose what is, as far as we can ascertain, the first deep learning solution that allows for multiple solutions.

Given a certain kinematic chain, one can readily obtain a training set consisting of pairs $(x,y)$ of end-effector positions $x$ and matching joint angles $y$, by sampling the latter and computing the former with forward kinematics. This straightforward way of obtaining the training set does not reflect the possibility of multiple solutions $y$, given the coordinates and orientation of $x$. In our framework, we employ a variational approach to the problem and sample, at inference time, from a distribution $P_x$ that is conditioned on the vector of Cartesian-space specifications $x$. 

Due to the structure of the kinematic chain, the IK problem can be seen as a hierarchical problem. Typically, the angle of the joint that is attached to the end-effector is uniquely determined by the location of all previous joints and the specifications in $x$. The previous joint may have multiple solutions given the joint angles that precede it. In general, as we move along the kinematic chain, from the fixed attachment point to the end effector, the number of possible configurations decreases. While this is true for any order in which we sequentially fix one joint, the kinematic chain is often equipped with a natural order, in which the first joints typically cause larger motions in Cartesian space.

In our model, this hierarchy is manifested by a sequential sampling of the joints from the distribution $P_x$. Namely, we parametrize $P_x$ as a sequential process in which the joints are sampled one by one and the sampling of each joint is conditioned on the values obtained for the previous joints.

The IK problem is often characterized by a discontinuous solution space. While for a given $y$, we can expect to see multiple solutions $y'$ that are close in the configuration space, there may be other solutions that obtain the desired position and orientation in $x$ using a completely different configuration. Our model addresses this by employing Gaussian Mixture Models (GMMs) during sequential sampling.   

The sequential sampling process, therefore, takes the following form. The distribution of the first joint is given as a GMM. The parameters of this GMM are computed by a neural network, given the desired position and orientation of the end effector $x$. The angle of the first joint $y^{(1)}$ (this is the joint that is the most distant from the end effector) is sampled by this GMM and the set of possible configurations for the remaining joints is reduced. A second GMM is then inferred in a way that is conditioned both on $x$ and on $y^{(1)}$. The second joint $y^{(2)}$ is sampled, and the process is repeated until all $N$ joints are obtained.

In the framework we propose, the parameters of each GMM are obtained by a neural network that receives the preceding joint locations as inputs. Conditioning on the input $x$ is obtained using a hypernetwork scheme, such that the parameters (weight matrices and biases) of the networks that provide the GMMs change dynamically, depending on $x$. This solution allows us to model the problem in a natural way, separating the conditioning on $x$ from the conditioning on the sampled values. 

Using the proposed neural IK solution, which we call IKNet, we present a path following method for recovering a sequence of joint location vectors given a sequence of smoothly varying end effector positions. This method runs online, such that at each time point the sampling of the joint angle depends on the preceding joints along the kinematic chain and on the angle of the same joint in the previous time step. The latter consideration ensures smoothness of the resulting path.

Our experiments demonstrate that IKNet outperforms a wide variety of IK methods, both optimization-based and learning-based. In the path following problem, our method generates multiple solutions, each more accurate and more stable than the single solution of the best baseline method. Additionally, we show that our probabilistic method displays robustness to noisy dimensions in the kinematic chain. Moreover, a relatively small number of examples is sufficient to finetune a trained model to perform well on a similar but unseen kinematic chain. Lastly, the representation learned by IKNet seems to help in learning other tasks.

\section{Related Work}

IK methods can be divided into analytical and numerical methods. Analytical methods~\cite{10.1115/1.2919218,diankov2010automated} provide a globally optimal solution, and in many situations multiple solutions, in an efficient and reliable way. However, the availability of analytical solutions is limited to models of limited complexity. Iterative (numerical) IK methods~\cite{buss2004introduction} update the vector of joint angular parameters through nonlinear optimization until convergence. A particular case is that of steerable needles, for which an optimization-based IK method was presented by \citet{4209361}.

Attempts to apply machine learning methods to IK include the application of One-Class SVM~\citep{scholkopf2002learning} by \citet{6094666}. \citet{973374} applied locally weighted projection regression~\citep{vijayakumar2000locally,klanke2008library} to this problem. In another work, \citet{de2008learning} have addressed IK with  Parametrized  Self Organizing Maps~\citep{walter1996rapid}.

A straightforward application of neural networks and other regression techniques to map between the vector $x$ of the end effector's position to the vector $y$ of joint locations~\cite{el2018comparative,duka2014neural}. Such methods fail to model the entire solution space for a given $x$ and are also, as we show empirically, less accurate. \cite{8211457} propose to heuristically divide the dataset to  reduce ambiguities. 

An important challenge for learning-based IK methods is to perform modeling online, with a given setup, and not based on a large training set~\citep{rolf2010bootstrapping,rolf2010goal,baranes2013active}. Such methods employ frameworks such as the one by \citet{moulin2014explauto}. In Sec.~\ref{sec:fewshot} we experiment with finetuning an existing model to model, with a relatively few samples, a model that deviates from it.

 IK-based learning methods have been applied for computer graphics purposes, in order to obtain a more natural motion~\cite{grochow2004style,10.1111:cgf.13089}. Learning in such cases is based on motion capture and other sources of data. In contrast to these methods, our method is aimed at finding all possible solutions for a given kinematic chain and the data we employ is synthetic data generated by this chain.

In order to perform sequential sampling, we employ a series of networks $g_k$, which are all conditioned on the pose vector $x$. To this end, we employ a hypernetwork~\cite{ha2016hypernetworks}. The hypernetwork scheme has two components: a primary network $g$, which outputs the computation result, and a hypernetwork $f$, which is used for conditioning on some input. The weights of network $g$ are not learned directly. Instead, they are provided as the output of network $f$. Therefore, the weights of $g$ are dynamic and vary based on the input of $f$. Hypernetworks have been used for RNNs since their inception, but we are not aware of any other application for a series of primary networks. 

While there exist alternative ways of conditioning, such as passing $x$ as an additional input to each  network $g_k$, hypernetworks provide a modular solution, in which the capacity of the conditioning network can be increased, while employing relatively shallow primary networks~\cite{galanti2020modularity}. This is useful in our IK framework, in which data is generated synthetically. In our experiments, we present a baseline that conditions on the end effector position without using a hypernetwork and demonstrate the advantage of employing hypernetwork-based conditioning in the context of IK. The hypernetwork structure that we employ performs hierarchical sampling. In the context of image generation, \citet{bensadoun2021meta} have combined a hypernetwork with a hierarchical sampling to obtain multiple valid solutions.

\begin{figure*}[t]
    \centering
    \includegraphics[trim={15cm 0 0 0}, clip, width=.96\linewidth]{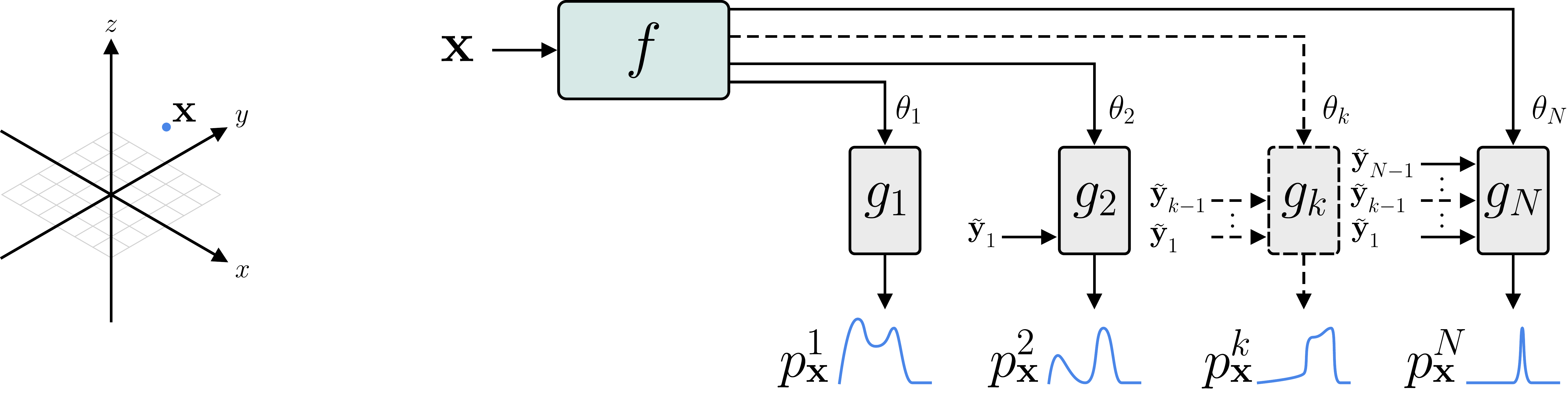}
    \caption{Illustration of the proposed method. Given a query input $\mathbf{x}$, which is the desired end-effector position, the hypernet $f$ maps its input to the set of parameters of $N$ primary networks $g_1,g_2, \dots,g_N$, where $N$ is the number of joints in the kinematic chain. The output $\Tilde{\mathbf{y}}_k$ of $g_k$ is a GMM $p_{\mathbf x}^k$, modeling the solutions distribution for joint $k$, and the input for each $g_k$ are all the previously sampled $\Tilde{y}_1, \dots, \Tilde{y}_{k-1}$.}
    \label{fig:main}
\end{figure*}

\section{Method}

In this section, we describe a method for learning the mapping from an end-effector position $\mathbf{x}$ to the distribution of the joint angles $P_x$, thereby enabling the sampling of joint angles $\Tilde{\mathbf{y}}\in\mathbb{R}^N\sim P_x$, such that applying $\Tilde{\mathbf{y}}$ to the $N$-joint kinematic model results in the end-effector at position $\mathbf{x}$.

At training time, we are given a set of pairs of vectors $\{(\mathbf{x}^{(i)},\mathbf{y}^{(i)})\}$, in which every vector $\mathbf{x}^{(i)}$ is matched with a single vector $\mathbf{y}^{(i)} \in S_x$, where $S_x$ is a set of possible matching y-space vectors for the vector $\mathbf{x}$. Note that our formulation allows for the existence of indices $i,j$ such that $\mathbf{x}^{(i)}=\mathbf{x}^{(j)}$, but $\mathbf{y}^{(i)}\neq \mathbf{y}^{(j)}$. To index the vector  $\mathbf{y}^{(i)}\in\mathbb{R}^N$, we use the superscript $\mathbf{y}^{(i)}$ to denote the $i$-th sample, and the subscript $\mathbf{y}_k$ to denote the $k$-th joint.

In the IK problem, $x$ is the position of the robot's end-effector, and $\mathbf{y}\in\mathbb{R}^N$ is the vector of joint angles, where $N$ is the number of joints in the kinematic chain. The end effector position can include its location or both its location and orientation. In the latter case, the number of plausible IK solutions decreases. 

Every valid Cartesian position $\mathbf{x}$ has one or more matching joint configurations $\mathbf{y}$, which collectively form the set $S_x$. Since the forward mapping is one to one, $S_x \cap S_x' = \O$ for $\mathbf{x}\neq \mathbf{x}'$. However, our method does not employ this fact.

Our goal is to learn to map every vector $\mathbf{x}$ to a conditional distribution $P_x$, such that the likelihood of every $\mathbf{y} \in S_x$ is high and, conversely, low for $\mathbf{y} \notin S_x$. 

Due to the hierarchical structure of the kinematic chain, we parametrize $P_x$ such that sampling a vector $\Tilde{\mathbf{y}}$ from this distribution is done sequentially 
\begin{align}
    \Tilde{\mathbf{y}}_1 &\sim p^1_x := p_x(\mathbf{y}_1) \\
    \Tilde{\mathbf{y}}_2 &\sim p^2_x := p_x(\mathbf{y}_2 | \Tilde{\mathbf{y}}_1) \\
    \dots \nonumber\\
    \Tilde{\mathbf{y}}_k &\sim p^k_x := p_x(\mathbf{y}_k | \Tilde{\mathbf{y}}_1,\dots, \Tilde{\mathbf{y}}_{k-1})\\
    \dots&~~\text{and} \nonumber\\
    P_x &:= \left [ p^1_x, p^2_x, \dots, p^N_x \right ].
\end{align}
Namely, the first element $\Tilde{\mathbf{y}}^1$ is sampled first from the distribution $p_x(\mathbf{y}^1)$, then the second element, $\Tilde{\mathbf{y}}^2$, in a way that is conditioned on the first, $\Tilde{\mathbf{y}}^1$, using the distribution $ p_x(\mathbf{y}_2 | \Tilde{\mathbf{y}}_1)$, and so on. This way of sampling is natural for kinematic chains, as mentioned in Sec.~\ref{sec:intro}.

Specifically, we model each part of the $P_x$ distribution $p^1_x,\dots, p^N_x$ as a Gaussian mixture model (GMM). This way, the distribution is able to capture sets $S_x$ that have a discontinuous shape, with multiple regions. For $m$ GMM components, the distribution $p^k_x$ is thus parameterized by a vector $\mathbf{m}_x^k \in \mathbb{R}^{3m}$ capturing the mean, variance, and mixture coefficient of each of the $m$ components.

Let $N$ be the dimensionality of the vectors $\mathbf{y}$. The method employs a hypernet structure, in which the hypernet $f$ maps its input $\mathbf{x}$ to the set of parameters of $N$ primary networks $g_1,g_2,\dots,g_N$. The mapping between $\mathbf{x}$ and $P_x$ and the sampling from this distribution is formulated as follows:
\begin{align}
    [\theta_1,\theta_2,\dots,\theta_N] & = f(x) \label{eq:hypernet1}\\
    \mathbf{m}_x^1 & = g_1(\theta_1) \label{eq:hypernet2}\\
    \Tilde{\mathbf{y}}_1 &\sim p_x^1 \label{eq:hypernet3}\\
    \mathbf{m}_x^2 &= g_2(\theta_2,\Tilde{\mathbf{y}}_1)\label{eq:hypernet4}\\
    \Tilde{\mathbf{y}}_2 &\sim p_x^2\label{eq:hypernet5}\\
    \mathbf{m}_x^3 &= g_3(\theta_3,\Tilde{\mathbf{y}}_1,\Tilde{\mathbf{y}}_2)\label{eq:hypernet6}\\
    \Tilde{\mathbf{y}}_3 &\sim p_x^3\label{eq:hypernet7}\\
    \dots \nonumber \\
    \mathbf{m}_x^N &= g_N(\theta_N,\Tilde{\mathbf{y}}_1,\Tilde{\mathbf{y}}_2,\dots,\Tilde{\mathbf{y}}_{N-1})\label{eq:hypernet8}\\
    \Tilde{\mathbf{y}}_N &\sim p_x^N\label{eq:hypernet9}
\end{align}
where for the primary networks $g_k$, the network parameters are mentioned explicitly as the first input parameter, and  $p_x^k$ is the GMM distribution with the parameters $\mathbf{m}_x^k$. Given an input sample $\mathbf{x}$, the hypernet $f$ returns, in Eq.~\ref{eq:hypernet1}, the parameters of the $N$ primary networks. Then, in Eq.~\ref{eq:hypernet2}, the GMM parameters of the distribution for the first element of $\Tilde{\mathbf{y}}$ are obtained through the primary network $g_1$. Subsequently, in Eq.~\ref{eq:hypernet3}, a value is obtained from this distribution. Conditioned on the sampled value, the parameters of the GMM of the second element in $\Tilde{\mathbf{y}}$ are obtained (Eq.~\ref{eq:hypernet4}), and this value is sampled (Eq.~\ref{eq:hypernet5}). The process continues until all $N$ values of the vector $\Tilde{\mathbf{y}}$ are obtained, each conditioned on the previous elements.

Fig.~\ref{fig:main} illustrates the proposed methods, where given a query input $\mathbf{x}$ for $f$, as a desired end-effector position, the network produced $N$ GMMs, as the number of joints, that we can sample from. Each sampled solution will be mapped to the end-effector position using forward kinematics.

\subsection{Training}
During training, we learn only the parameters of the hypernetwork $f$. The parameters of the primary networks $g_k$ are given by $f$ and change based on the input to this network. In the training procedure, we employ a teacher forcing scheme, in which the values of the training sample $\mathbf{y}^{(i)}$ are employed instead of sampling.

The loss term we employ during training maximizes the log likelihood of the training samples
\begin{align}
\label{eq:L1}
    \mathcal{L} &= - \sum_i \log \prod_{k=1}^{N} p^k_{\mathbf x^{(i)}}\left ({\mathbf{y}}^{(i)}_k|\mathbf{y}_1^{(i)},\dots,\mathbf{y}_{k-1}^{(i)}\right )
\end{align}
where $\mathbf{y}^{(i)}_1, \dots, \mathbf{y}^{(i)}_{k-1}$ are the ground truth values, \ie,
no sampling takes place, and instead of sequential sampling as in Eq.~\ref{eq:hypernet3},\ref{eq:hypernet5},\ref{eq:hypernet7},~and \ref{eq:hypernet9}, the values of $\mathbf{y}^{(i)}$ are used. The training sample $\mathbf{x}^{(i)}$ is manifested through the distributions $p^k_{\mathbf x^{(i)}}(\mathbf{y}_k|\dots)$ , which are based on the GMM parameters $\mathbf{m}_{\mathbf x^{(i)}}^k$ (Eq.~\ref{eq:hypernet2},\ref{eq:hypernet4},\ref{eq:hypernet6},~and \ref{eq:hypernet8}).

\subsection{Architecture}
The network $f$ is a 4-layer fully-connected network. Each linear layer has a dimension of 1024, with ReLU and batch-norm following each layer. The last layer of $f$ is followed by $N$ projection layers that map the last dimension of 1024 to the vector of weights, $\theta_k$, for each network $g_k$.

The networks $g_k$ take as an input the weights produced by $f$ and a sequence of joint angles $\mathbf{y}_1,\dots,\mathbf{y}_{k-1}$. 
Each network is composed of three linear layers with a hidden dimension of 256, and ReLU activation between the layers. The output is a vector of $3m$ elements, where a subset of $m$ elements denote the prior for each GMM component. 

In order not to explicitly select an optimal value for the parameter $m$, we set it at a very high value of $m=50$, and make sure that the vector of priors is sparse. Specifically, the relevant (\ie, not mean- or variance-related) $m$ values produced by each $g_k$ undergo a sparsemax~\cite{martins2016softmax} operation.

\section{Path Following}
\label{sec:pf}

As an application of the IK network, we present a method for recovering a sequence of joint locations $\mathbf{Y}=[\mathbf y^{1},\mathbf y^{2},\dots,\mathbf y^{n}]$, given a desired path of end-effector locations $\mathbf{X}=[\mathbf x^{1},\mathbf x^{2},\dots,\mathbf x^{n}]$. 
Ideally, given a many-to-one situation, one would like to obtain multiple different sequences, each of which should depict a smooth path in the joint location space, which matches (by applying forward kinematics) the desired sequence $\mathbf{X}$.

To achieve this, we employ a path following method. At each point along the desired path, we have the joint angles in the previous time-step $\bar{\mathbf{y}}^{t-1}$, and the desired end-effector position $\mathbf x^t$. By employing our network ($f$ then $g_1$), we obtain a GMM distribution $p_{\mathbf x^t}^1$ for the first joint at time $t$. 

To maintain a smooth transition, we sample $\bar{\mathbf{y}}^t_1$ (\ie, the first joint of the current time step) from $p_{x^t}^1$, subjected to the neighborhood $\Omega$ of $\bar{\mathbf y}^{t-1}_1$ of radius $r$, which we denote by $\Omega({\bar y^{t-1}_1}, r)$:
\begin{align}
    \bar{\mathbf{y}}^{t}_1 \sim p_{\mathbf x^t}^1|_{\Omega({\bar{\mathbf{y}}^{t-1}_1},r)}
\end{align}
This way, we maximize the smoothness of the path, while sampling from the learned distribution of the joint location that is conditioned on the end effector position in the current time-step, Thus emphasizing GMM components that are closer to the the joint angles in the previous time-step.

We now have the angle of the first joint at time $t$, denoted by $\bar{\mathbf{y}}^t_1$. We repeat the process using the same path following procedure, this time applied to $p_{\mathbf x^t}^2 = g(\theta_2,\bar{\mathbf{y}}^t_1)$. After this, the process is iterated for the rest of the joints. In our experiments we choose $r = 0.1$ radian.

\section{Experiments}

Our experiments check the performance of the IK method for isolated poses as well as for paths. We also present an experiment evaluating the representation learned by IKNet.

\paragraph{Metrics} We compare the accuracy of the results using the following scores: (1) the mean Euclidean distance over the sample between $FK(\bar{\mathbf{y}})$ and $\mathbf x$, where $\bar{\mathbf{y}}$ is the obtained solution and $FK$ stands for the forward kinematics, (2) accuracy, which is the ratio of solutions for which the Euclidean distance is less than a set threshold.

\begin{table}[t]
    \centering
    \caption{Results for the two 2D chains.}
    \label{tab:2D}
    \vskip 0.15in
    \begin{tabular*}{\linewidth}{l@{\extracolsep{\fill}}cc}
    \toprule
    & Mean distance (cm) & Accuracy\\
    \midrule
    2-joints & 0.6 $\pm$ 0.6 & 97.6\% $\pm$ 4.7\%\\
    4-joints & 0.8 $\pm$ 0.6 & 95.6\% $\pm$ 3.7\%\\
    \bottomrule
    \end{tabular*}
\end{table}
\subsection{Evaluation on a 2D chain}
In order to illustrate the ability of our method to capture the distribution of solutions, we train and evaluate with two settings of 2-dimensional chains -- two joints (2J) and four joints (4J), where each joint is limited to $180$ degrees for each direction. We train our model on a dataset of 20K random (reachable) points, and test on a different set of 1K (reachable) points. Tab.~\ref{tab:2D} show results for mean distance error and accuracy, where we measure accuracy as the percentage of points that are up to 2cm from the end-effector.

In the case of the 2J-chain, we notice, analytically, that the number of solutions for the first and second joints are 2 and 1, respectively. In the case of the 4J-chain, there is an infinite number of solutions for a given end-effector, since the kinematic chain has more degrees of freedom than the end-effector.

\begin{figure*}[t]
    \setlength{\tabcolsep}{1pt} 
    \renewcommand{\arraystretch}{1} 
        \vskip 0.15in
    \centering
    \begin{tabular}{cccc}
    \includegraphics[width=0.25\linewidth]{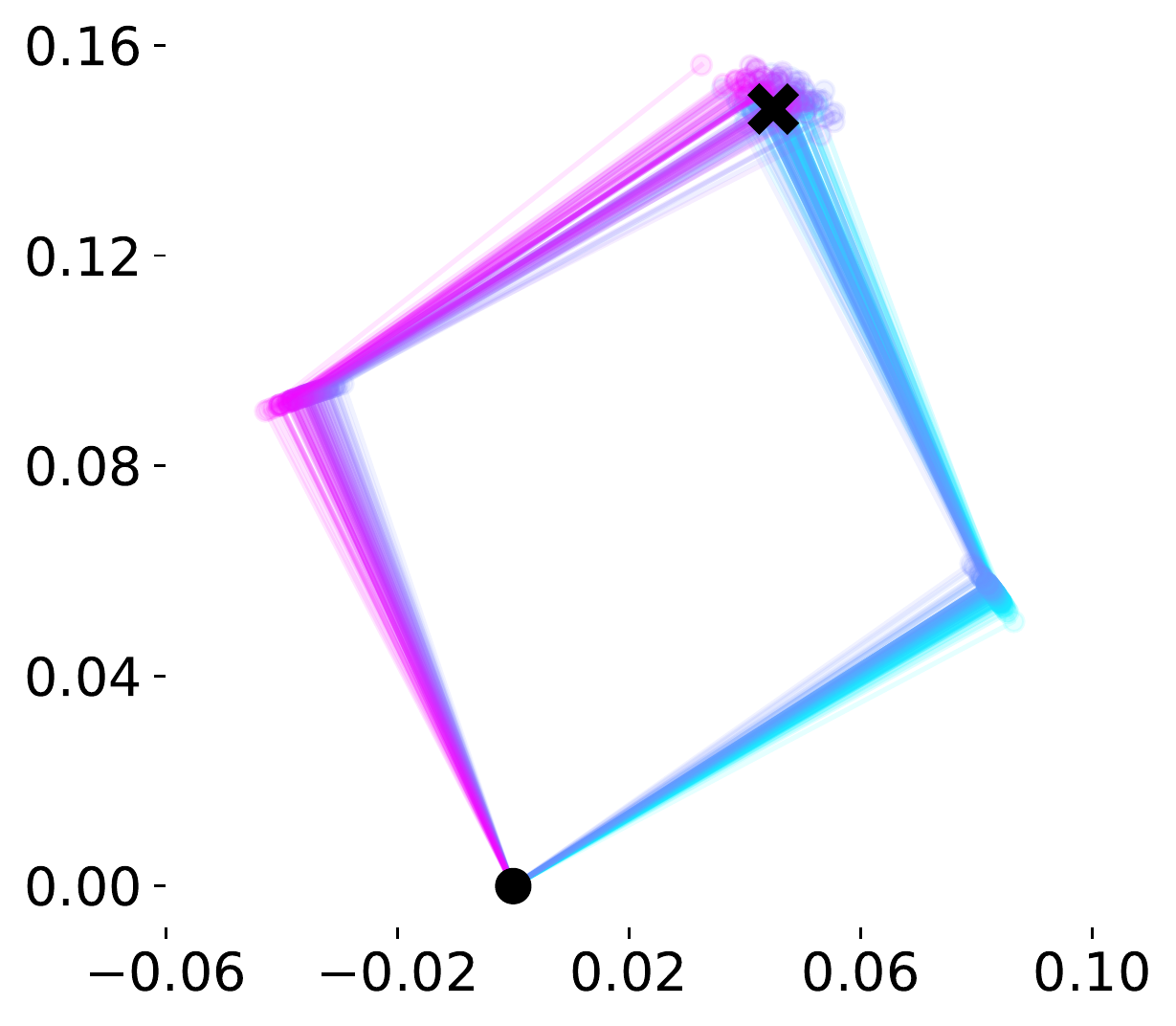} &
    \includegraphics[width=0.25\linewidth]{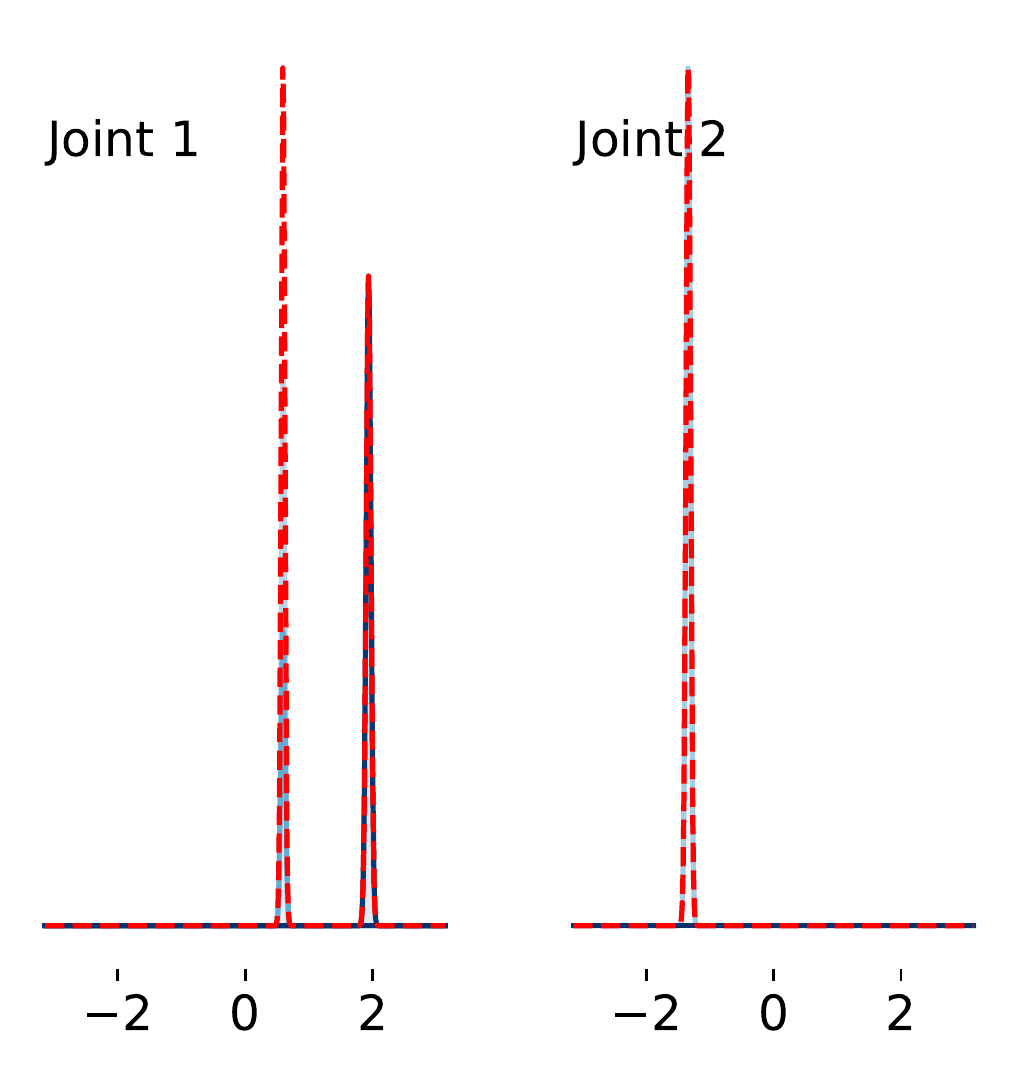}&
    \includegraphics[width=0.25\linewidth]{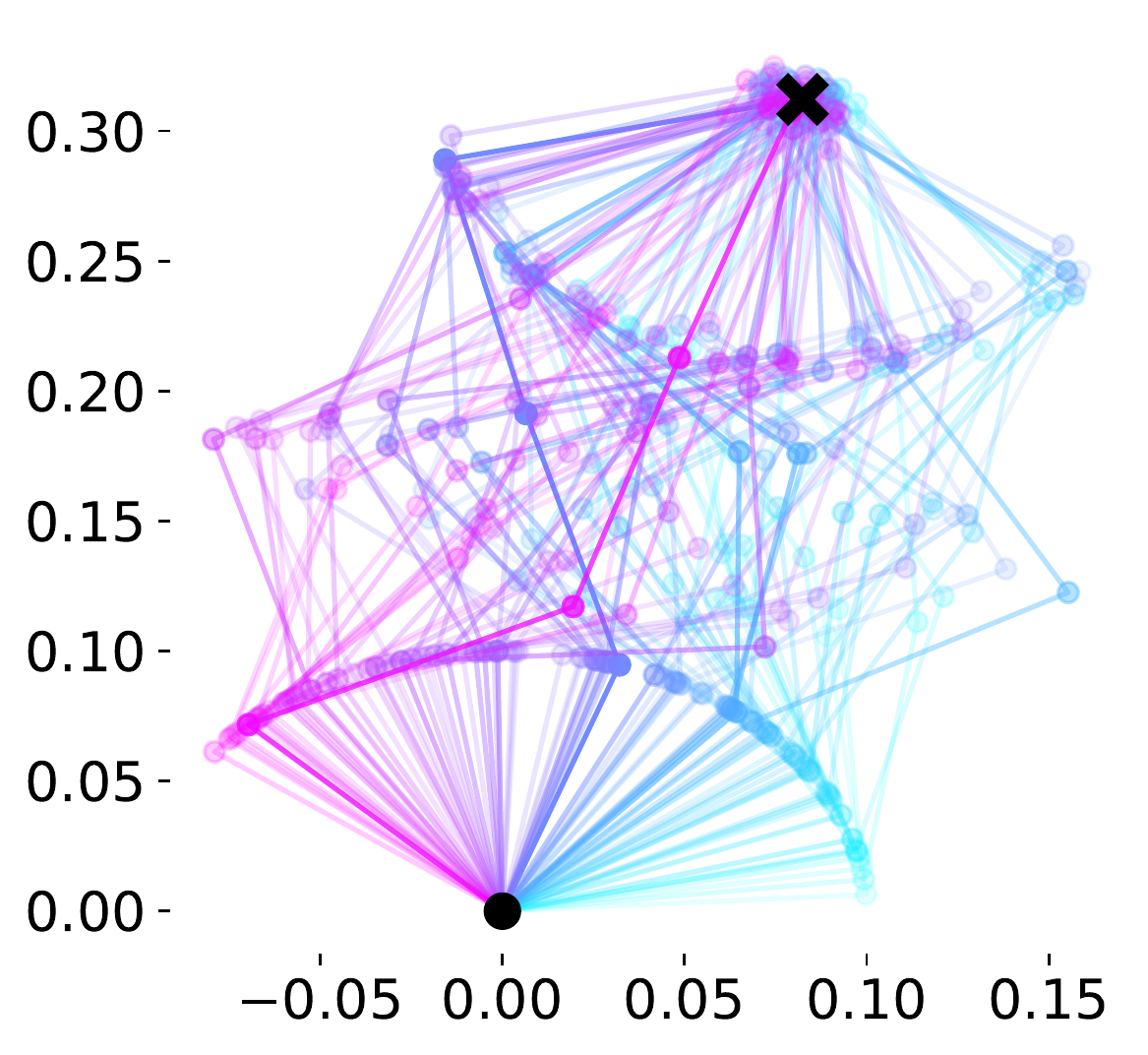} &
    \includegraphics[width=0.25\linewidth]{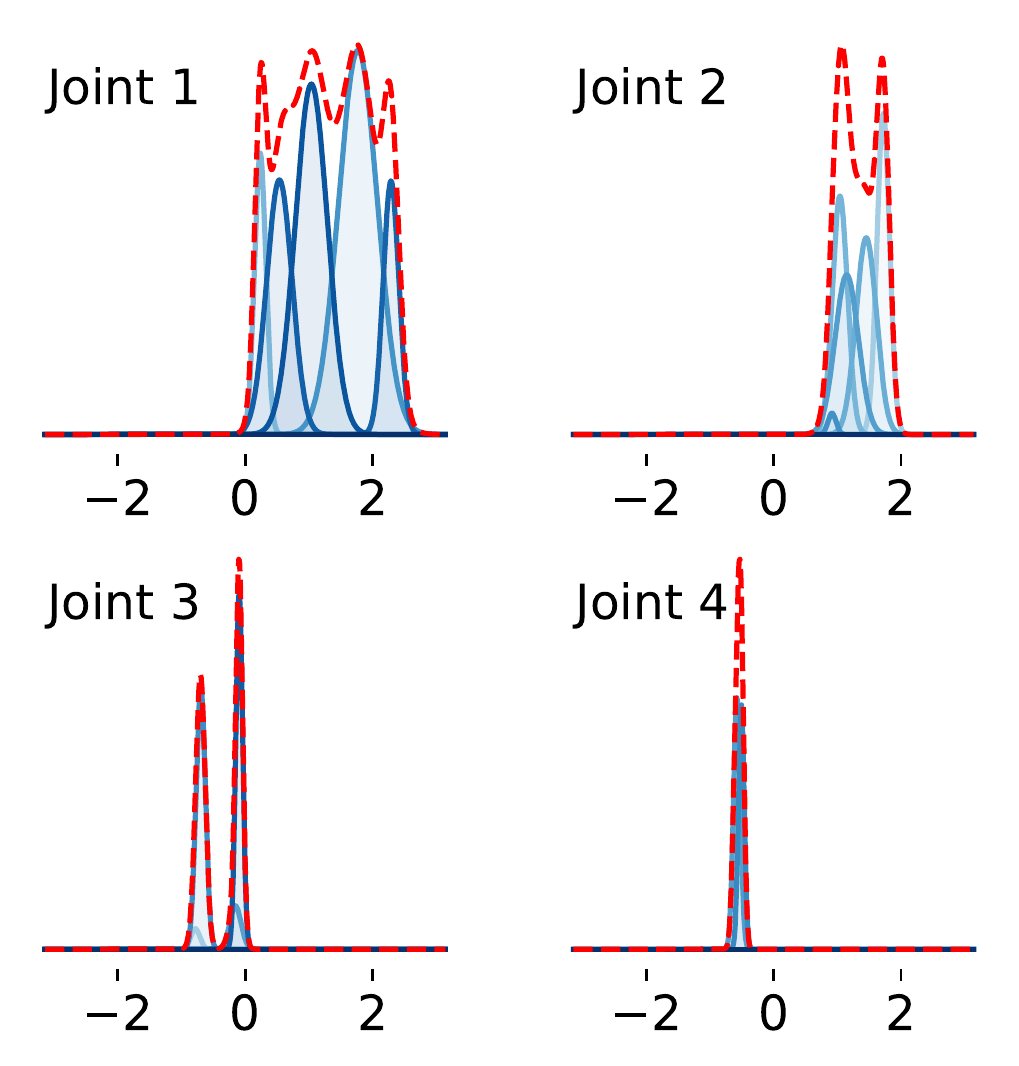}\\
    (a) & (b) & (c) & (d)
    \end{tabular}
\caption{Illustration of the two 2D chains settings. We present 100 solutions for each chain and for a given end-effector position. (b,d) show the learned GMMs for each joint, and (a,c) show the sampled chain layouts, where opacity represents the probability of the layout, and starting / end-effector positions are illustrated with circle and X symbols, respectively.}
    \label{fig:chains}
\end{figure*}

In Fig.~\ref{fig:chains} we illustrate 100-sampled solutions for each chain, and for a given end-effector position. 
Fig.~\ref{fig:chains}(b,d) show the learned GMMs for each joint, and Fig.~\ref{fig:chains}(a,c) show the sampled chain layouts, where opacity represents the probability of the layout, and starting / end-effector positions are illustrated with circle and X symbols, respectively.

As can be seen, the network learns to model the two layout solutions for a given point in the 2J-chain. The GMM of the first joint is collapsed into two main means, and the GMM of the second joint is collapsed into a single mean, both with low variance. In the 4J-chain, we can observe that the first joint's angles are spread according to its GMM distribution, while the third and fourth joints collapsed to the same distribution of results as in the 2J-chain.

\subsection{Comparison with IK methods}
In the following section, we present the experiment setting used for evaluating our method, termed IKNet, against well-established IK methods, as well as machine-learning baselines. We use three different robotic arms, with different levels of complexity, as our benchmark kinematic chains.

\subsubsection{Baselines}
\paragraph{Numerical methods} We experiment with three types of optimization-based methods, (i) the Damped least squares Jacobian~\cite{wampler1986manipulator} -- which optimizes the end-effector position using the Jacobian of the model, (ii) the IK software package IKPy\cite{ikpy} -- which optimizes the position using `L-BFGS-B'~\cite{zhu1997algorithm} optimizer, and (iii) a differentiable model of the forward-kinematics package, DiffNEA, by~\citet{pmlr-v120-sutanto20a} -- which optimizes the joint angle to minimize the L2 distance between the current and desired end-effector position. 

Each method was initialized with multiple starting points in order to obtain multiple solutions, and ran for the same amount of time per sample during the optimization step for a fair comparison.

\paragraph{Learning methods} We construct three different network models that capture different aspects of our method. First, we build an MLP network with depth 3 and width 1024, which takes as an input the desired end-effector, and outputs all joint angles at once. This baseline is incapable of generating multiple solutions. 
Second, instead of modeling conditional distributions sequentially, we use a network $f$ with the same architecture as our hypernetwork to output mean vectors, covariance matrices (via Cholesky decomposition) and selection weights, to model the distribution as a mixture of multivariate Normal random variables, sampling all joints at once. 

Last, we experiment with a recurrent neural network (RNN) architecture for modeling the sequential properties of the distribution of joint solutions. The RNN architecture is composed of a shared weights GRU~\cite{cho2014properties} module for all joints, and an independent MLP part for each joint before and after the GRU module. This is done in order to model the unique part of each joint. To reflect the angles of the preceding joints, while reducing the repetition in the solution space, we project the end-effector position to the current coordinate system, reflecting the location of the joint after the preceding joints have determined its location. 

\begin{figure}[t]
    \centering
    \includegraphics[width=\linewidth]{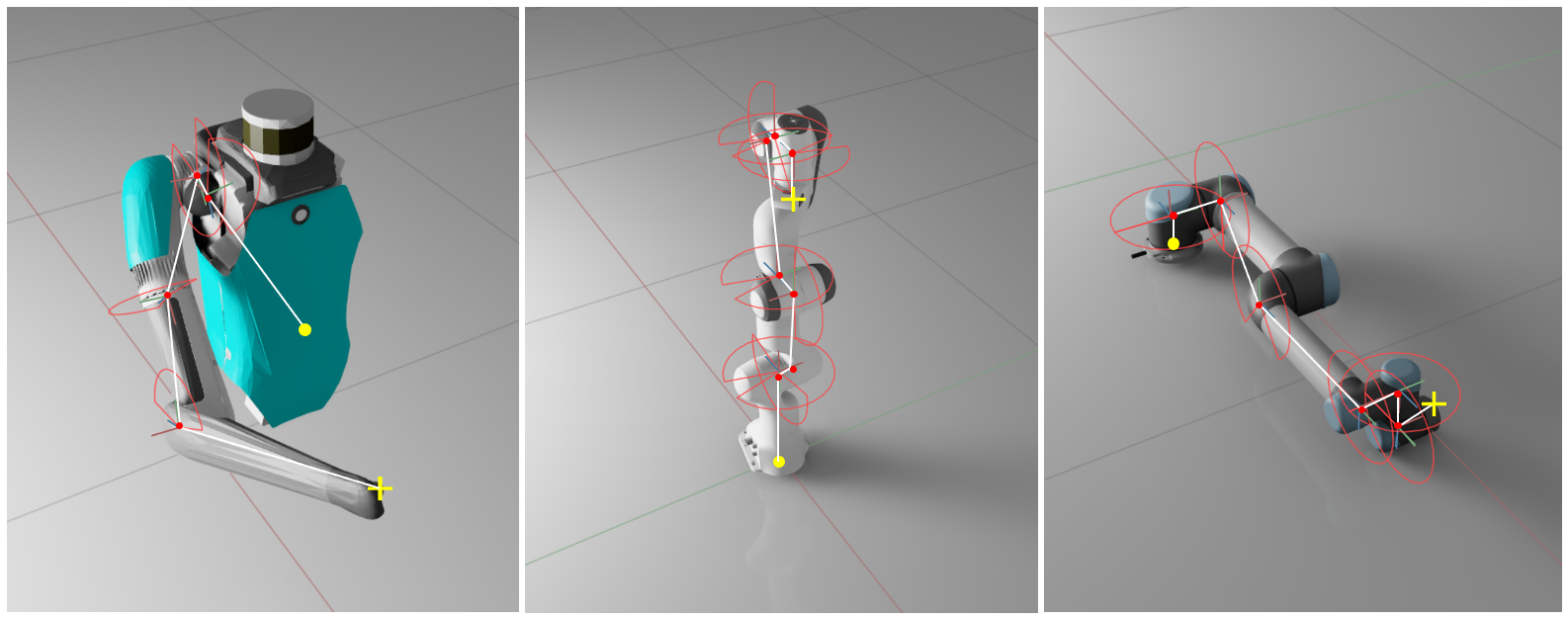}
    \caption{Illustration of the three kinematic chains used in our experiments. From left to right, Digit arm as obtained from the DigitRobot.jl repository \url{https://github.com/adubredu/DigitRobot.jl} -- containing 4 DoFs, Franka panda -- containing 7 DoFs and UR5 -- containing 6 DoFs. The joints axis are illustrated in red.}
    \label{fig:robots}
\end{figure}
\subsubsection{Kinematic Chains (datasets)}
We demonstrate our method on 3 different kinematic chains, which differ in their scale and degrees of freedom (DoFs) (Fig.~\ref{fig:robots}). All demonstrated kinematic chains are redundant, i.e the number of DoFs is greater than the task space dimension (for the $3D$ inverse position kinematic task employed in the experiments, this dimension is 3). 

\paragraph{Digit arm} Digit by Agility robotics is a humanoid, bipedal robot, made for work in environments designed for humans. Digit's upper torso is integrated with two 4-DoFs arms aimed for basic manipulation and object carrying tasks. As a source for this model, we employs the public repository by Alphonsus Adu-Bredu that is available at \url{https://github.com/adubredu/DigitRobot.jl}. According to this resource, Digit's arm DoF angular ranges are [(-1.3, 1.3),(-2.5, 2.5),(-1.75,1.75),(-1.35, 1.35)] radians.

\paragraph{UR5} UR5 by Univeral Robots is an industrial, flexible, lightweight, 6-DoFs robotic arm with working radius of up to $85.0cm$. The DoFs angular range are [(-3.14, 3.14),(-3.14, 3.14),(-2.5, 2.5),(-3.14, 3.14),(-3.14, 3.14),(-3.14, 3.14)] radians.

\paragraph{Franka} Franka Emika Panda is a 7-DoFs programmable robotic arm with working radius of up to $85.5cm$. The DoF angular ranges are [(-2.9, 2.9),(-1.76, 1.76),(-2.9, 2.9),(-3.07, -0.07),(-2.9, 2.9),(-0.02, 3.75),(-2.9, 2.9)] radians.

\subsubsection{Results}

The results are presented in Tab.~\ref{tab:main_table} and depict mean and Standard Deviation on a test set of reachable arm locations. As can be seen, IKNet outperforms both the optimization based baselines and the learning based baselines is terms of mean distance, with the exception of IKPy outperforming somewhat better than our method on average, but with much higher variance. 

Accuracy is measured at a 10cm threshold. Our method outperforms all baselines in this metric. Lastly, we measure the actual runtime per method and report results for 100 executions. For this purpose, the learning based methods were run on a CPU. As can be seen, our method is considerably more efficient than the optimization based methods. While the runtime cannot be taken at face value, since it is implementation-dependent, our method has an inherent advantage since it does not iterative. 

\begin{table*}[t]
    \centering
    \caption{Inverse kinematic results on three kinematic chains. Our results are provided as the average of 100 samples and not based on the most likely solution, which would improve IKNet results. For each method we present the mean across the test set as well as the Standard Deviation. As can be seen, our method achieves best accuracy and standard deviation across all robots, and best accuracy for Digit arm and UR5. For Franka panda, our method is compatible with the results of IKPy, but with better standard deviation. }
    \medskip
    \centering
    \begin{tabular}{llll}
    \toprule
    & Distance (cm) & Accuracy & Runtime (s) \\
    \midrule
    \multicolumn{4}{c}{DIGIT - Arm (4 DoFs)}\\
    \midrule
    Damped least squares Jacobian & 8.3 $\pm$ 19.6 & 80.4\% $\pm$ 15.9\% & 0.0400\\\
    IKPy & {7.5 $\pm$ 16.0} & {79.0\% $\pm$ 13.6\%} & 0.0850\\
    DiffNEA & 28.7 $\pm$ 20.9 & 21.8\% $\pm$ 6.5\% & 0.1100\\[-\jot]
 \multicolumn{4}{@{}c@{}}{\makebox[.7\linewidth]{\dashrule[black!40]}} \\[-\jot]
 MLP & 12.3 $\pm$ 1.2 & 56.2\% & 0.0002 \\
    RNN & 12.7 $\pm$ 11.0 & 53.5\% $\pm$ 24.2\% &  0.0090 \\
    Multivariate GMMs & 4.4 $\pm$ 3.7 & 92.3\% $\pm$ 12.7\% & 0.0020\\[-\jot]
     \multicolumn{4}{@{}c@{}}{\makebox[.7\linewidth]{\dashrule[black!40]}} \\[-\jot]
    IKNet & \textbf{2.3 $\pm$ 1.7} & \textbf{99.5\% $\pm$ 1.3\%} & 0.0070\\
    
    \midrule
    \multicolumn{4}{c}{UR5 (6 DoFs)}\\
    \midrule
    Damped least squares Jacobian & 10.9 $\pm$ 22.3 & 76.2\% $\pm$ 40.9\% & 0.2000\\
    IKPy & {6.5 $\pm$ 15.4} & {81.7\% $\pm$ 9.9\%} & 0.1200\\
    DiffNEA & 36.4 $\pm$ 24.5 & 15.3\% $\pm$ 4.8\% & 0.1100\\
    [-\jot]
     \multicolumn{4}{@{}c@{}}{\makebox[.7\linewidth]{\dashrule[black!40]}} \\[-\jot]
    MLP & 59.1 $\pm$ 0.9 & 2.6\% & 0.0002\\
    RNN & 5.7 $\pm$ 8.9 & 84.7\%  $\pm$ 7.4\% &  0.0090\\
    Multivariate GMMs & 5.0 $\pm$ 3.8 & 89.8\% $\pm$ 15.5\% & 0.0050\\[-\jot]
     \multicolumn{4}{@{}c@{}}{\makebox[.7\linewidth]{\dashrule[black!40]}} \\[-\jot]
    IKNet & \textbf{2.8 $\pm$ 2.1} & \textbf{98.8\% $\pm$ 1.9\%} & 0.0120\\
    
    \midrule
    \multicolumn{4}{c}{Franka (7 DoFs)}\\
    \midrule
    Damped least squares Jacobian & 4.4 $\pm$ 13.3 & 88.7\% $\pm$ 6.8\% & 0.1100\\
    IKPy & \textbf{2.1} $\pm$ 7.6 & {91.2\% $\pm$ 6.3\%} & 0.1650\\
    DiffNEA & 22.8 $\pm$ 18.4 & 29.4\% $\pm$ 8.9\% & 0.1100\\[-\jot]
     \multicolumn{4}{@{}c@{}}{\makebox[.7\linewidth]{\dashrule[black!40]}} \\[-\jot]
    MLP & 57.7 $\pm$ 1.12 & 5.4\% &  0.0003 \\
    RNN & 4.9 $\pm$ 6.1 & 84.7\% $\pm$ 7.4\% &  0.0200\\
  Multivariate GMMs & 6.5 $\pm$ 4.6 & 82.1\% $\pm$ 16.8\% & 0.0100\\[-\jot]
   \multicolumn{4}{@{}c@{}}{\makebox[.7\linewidth]{\dashrule[black!40]}} \\[-\jot]
    IKNet & {3.1} $\pm$ \textbf{2.6} & \textbf{98.0\% $\pm$ 2.0\%} & 0.0300 \\
    \bottomrule
    \end{tabular}
    \label{tab:main_table}
\end{table*}

\begin{figure*}[t]
    \centering
    \begin{tabular}{c}
    \includegraphics[width=\linewidth,trim={0 0 180 0}, clip]{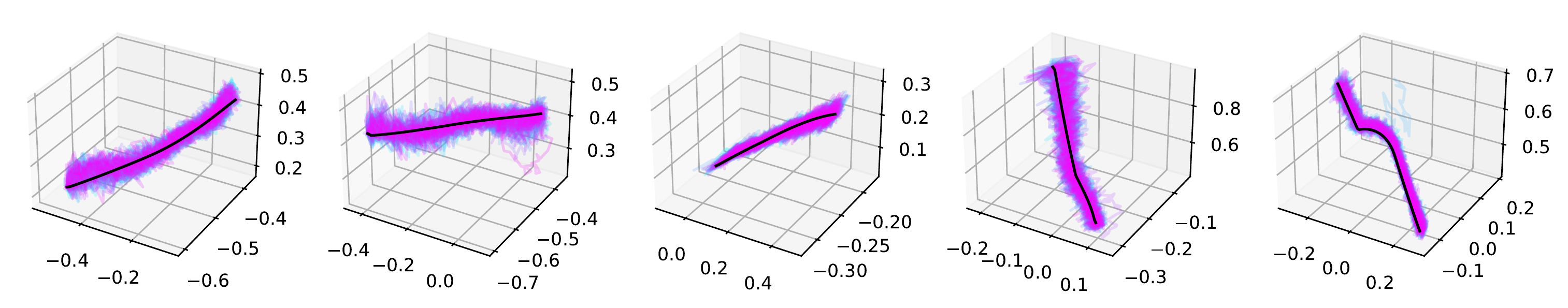}\\
    \includegraphics[width=\linewidth,trim={0 0 180 0}, clip]{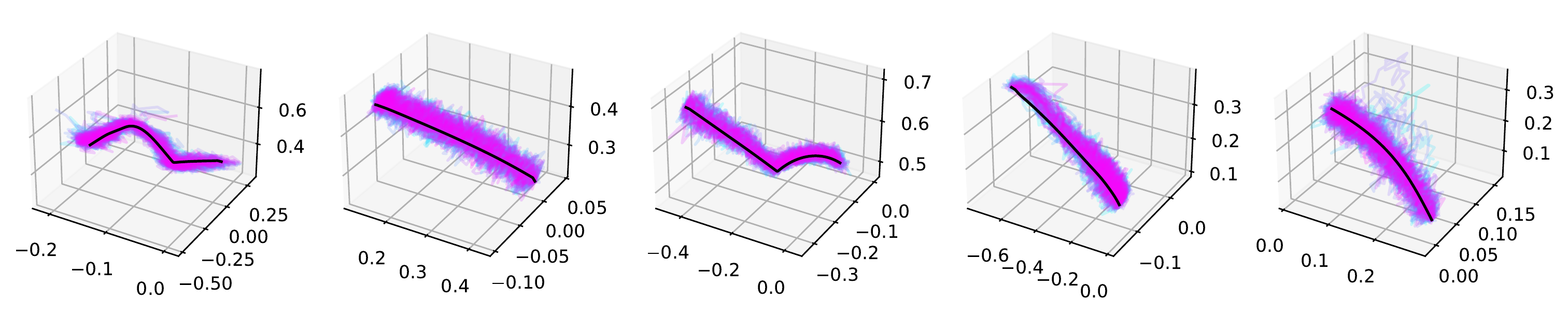}\\
    \end{tabular}
    \caption{Path following. Given a sequence of end effector locations, we use a path following method in order to recover multiple possible trajectories. The axis of the plots represent the angles along the first three joints of the Digit arm.}
    \label{fig:traj}
    \vspace{-0.2cm}
\end{figure*}
\subsection{Path following}

We next evaluate the path following methods presented in Sec.~\ref{sec:pf}. This experiment does not involve any training and evaluation sequences ${\mathbf{X}}$ of end effector positions were generated by moving the Digit arm smoothly. Each sequence is of length 50.

As a baseline to the path following method we propose, we employ IKPy, which is the best baseline method found in our single-position experiments. As an iterative method, IKPy finds a solution that is close to the starting point and can, therefore, be applied sequentially to the positions along the end effector's path to obtain a smooth trajectory in the joint space.

Comparing the Euclidean error in the end effector positions, the paths we generate obtain a similar average error to that of IKPy (3.0cm). However, for our method this is an average over 100 different paths and not the results for the best path, while for IKPy it is the result for the single path that starts at the ground truth joint position and varies smoothly (an ideal setting for IKPy). When selecting the generated path with the highest fidelity among all 100 generated paths (still considerably faster than running IKPy), we obtain an average error of 1.8 cm.

Fig.~\ref{fig:traj} presents the multiple trajectories obtained per a single target sequence $\mathbf{X}$. As can be seen, the generated paths present a high degree of variability.

\subsection{Robustness and few-shot learning}
\label{sec:fewshot}
Since our model is learning-based and since it employs a sampling-based approach, it can naturally model noisy robot dimensions. To demonstrate this we created a set
of 55 4-jointed 3D arms that differ by at most ±20\% in the length of each segment of the chain. We then learn one IKNet per arm and one based on the entire data. At test time, we sample 10 new random arms and evaluate the 55 single arm models and the one trained on the entire dataset. 

Out of the 55 random models, some are more similar than others and are ranked by the accuracy (the error ranges between 2.1-3.6cm, on average). The model that we train on the entire data is, on average, at the 75th
percentile of the results (standard deviation 6\%), which is a clear indication that the unified model learns a robust solution that matches many random arms.

Additionally, we expect our method to have an advantage in scenarios in which one needs to learn to perform IK from a few real measurements. To test this, we propose to leverage transfer learning in order to tackle a very common real-world scenario, in which a given robot deviates from the specifications.

We learn an IK model $\mathcal{M_{A}}$ for a 3D kinematic chain with 4 joints (“arm A”) based on 100K
samples. Then, we sample a new arm (“arm B”) in which the segment lengths vary randomly by up to ±20\% from the original design. We then fine-tune $
\mathcal{M_{A}}$ based on samples from arm B, obtaining model $
\mathcal{M_{B}}$ . 

\begin{figure}[b]
    \centering
    \includegraphics[width=\linewidth]{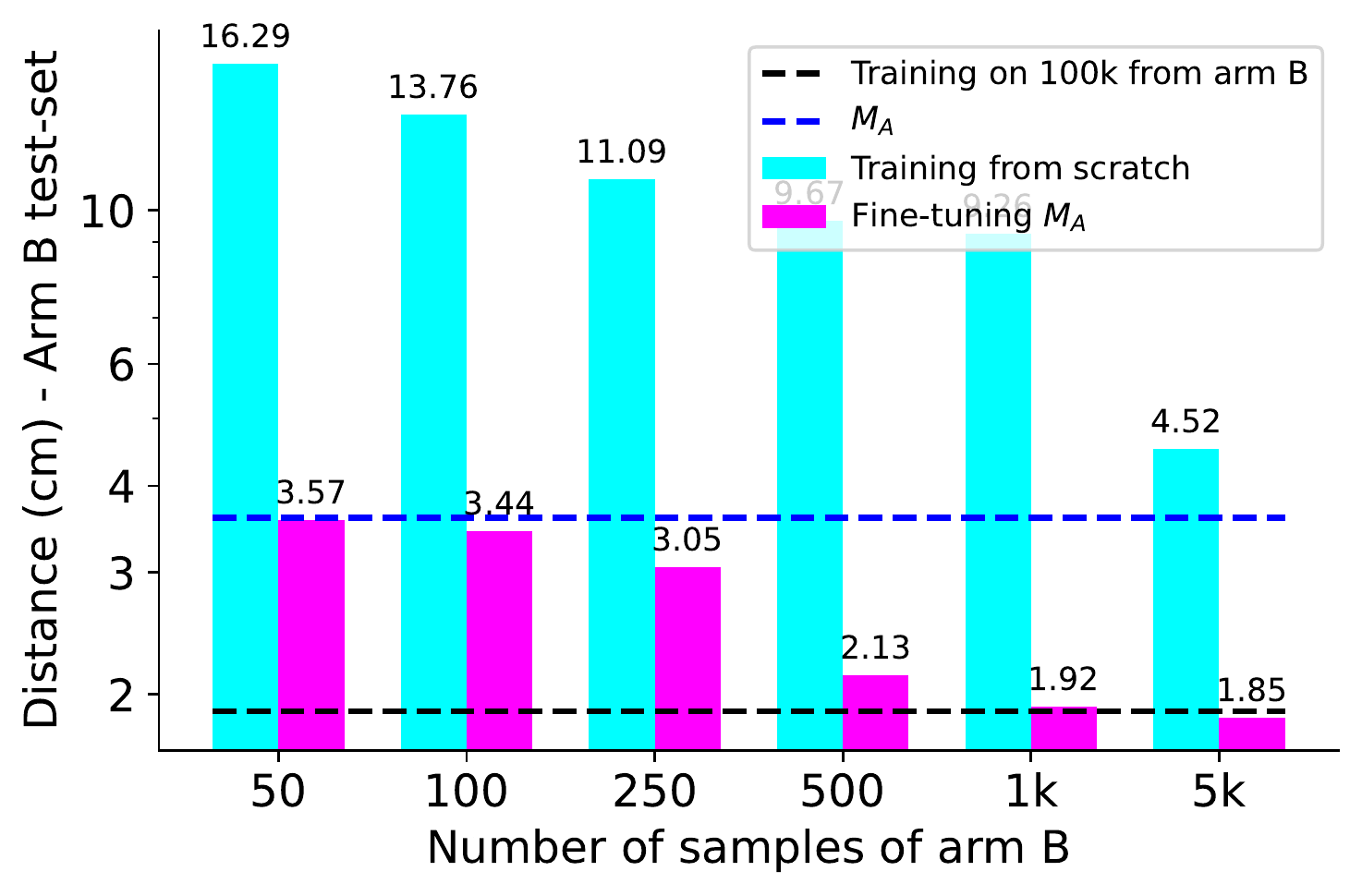}
    \vspace{-9mm}
    \caption{The number of training points during the training (magenta) or fine-tuning (cyan) phase of arm B vs.~the forward kinematics error (cm) on the test-set of arm B.}
    \label{fig:exp3graph}
\end{figure}

The results are presented in Fig.~\ref{fig:exp3graph}. Applying $\mathcal{M_{A}}$ on the test set of arm B yields an error of 3.62cm (dashed blue line). Training on 100K samples for arm B yields 1.84cm (dashed blackline). With 1K samples, the finetuned model $
\mathcal{M_{B}}$ obtains a similar error of 1.92cm, while a model trained from scratch with the same number of samples has a much higher error
of 9.26cm. Evidently, with a relatively small number of training points, one
reaches the same level of accuracy that can be obtained from 100K samples of arm B.

\subsection{Learning Seq2Seq mapping}

Lastly, in order to demonstrate that the representation learned by IKNet is powerful, we consider the task of learning to map a single sequence of effector poses $\mathbf{X}=[x^{1},x^{2},\dots,x^{100}]$ to a sequence of joint angles $\mathbf{Y}=[y^{1},y^{2},\dots,y^{100}]$. For this seq2seq task, we employ a Transformer~\cite{vaswani2017attention}. 

The transformer is trained on 10k pairs of matching sequences, each of length 100, from the UR5 chain. We compare two variants.  In the first, the transformer is trained to map $\mathbf{X}$ to $\mathbf{Y}$ directly. In the second, we employ representation obtained by $f'$, which is the mapping between an input of network $x$ and the activations of the last layer of $f$, before the $N$ projections to the weights of the networks $g_k$. In this case, the sequence to sequence problem learned is between $f'(\mathbf{X})= [f(x^{1}),f(x^{2}),\dots,f(x^{100})]$ and $\mathbf{X}$.

We employ a 6-layers deep Transformer encoder, with 8 attention heads. We also use an embedding of size 128, as our hypernetwork embedding size.  When training with raw inputs, a trainable linear layer projecting the 3-dimensional input to 128 dimensions is applied, in order to allow for fair comparison.

\begin{figure}
    \centering
    \includegraphics[trim={0 0 0 0}, clip, width=.98\linewidth]{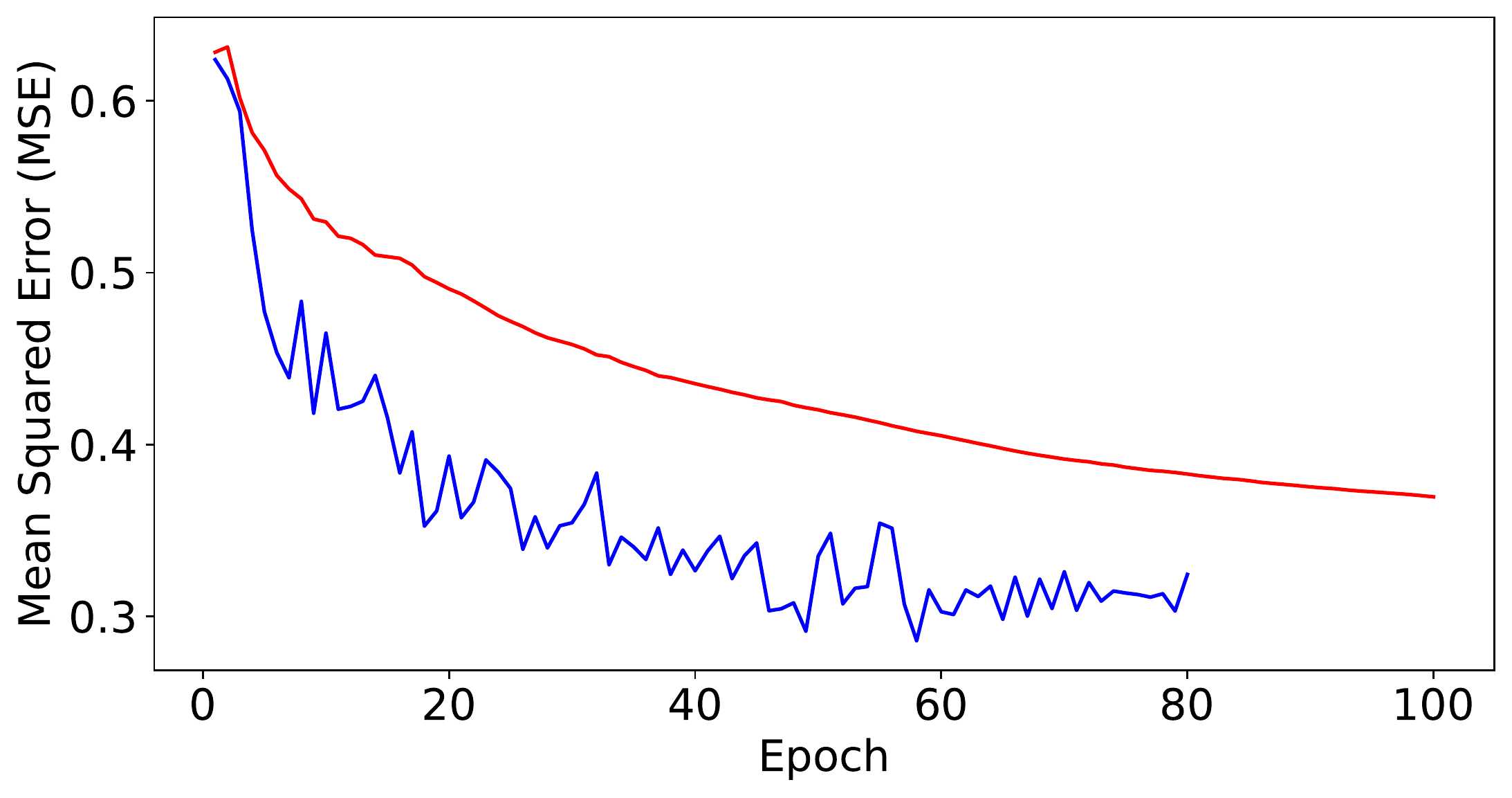}
    \caption{Convergence plots when training a sequence to sequence transformer model based either on the sequence of the end effector positions (red) or its embedding by network $f$ (blue).}
    \label{fig:transformers}
\end{figure}

In order to evaluate the results, we employ a test set of $100$ trajectories of $(\mathbf{X},\mathbf{Y})$, each of length 100. The convergence graphs for both methods are presented in Fig.~\ref{fig:transformers}. The plots depict the mean squared Euclidean error on the test set per epoch. We stopped training the  transformer model with the IKNet representation after 80 epochs, since it converged faster. Therefore, the two plots are of different lengths. Evidently, employing the IKNet representation leads to faster convergence and to an overall lower error.

\section{Discussion}

The most successful baseline methods are designed to provide one IK solution, given an initial position from which optimization starts. In contrast, variability is natural in our model, which is a feed-forward model and not an iterative optimization model.

The runtime of the method is a good indicator of its applicability. Reductio ad absurdum, without any limit on the runtime or the number of restarts, almost all optimization methods would reach optimal results and fully characterize the solution space. In the experiments, we relied on the default parameters provided in the implementation of each method. In the same vain, it is possible to use an optimization method on top the solutions provided by IKnet and obtain negligible MSEs. We avoid this in order to provide the raw results as returned by this neural model.

The more degrees of freedom the kinematic chain has, the easier it is for an optimization method to find a single valid vector of joint angles, given the end effector's location. However, modeling the ambiguity, which is the task solved by our network, becomes more difficult. In the Digit arm and UR5 experiment, where there are 4 and 6 DOF, respectively, our advantage over the baselines is more pronounced than it is for Franka with 7 DOF. 

Characterization of the entire probability distribution can also help achieve what is called robust inverse kinematics~\cite{sinha2019computing}. In this setting, one would like to select the IK solution that is the most stable with respect to errors in the joint angles. A robust solution is one such that the entire ball in joint angle space (where $y$ resides) leads to end effector positions within a certain tolerance of the desired position and angle ($x$).

\section{Conclusions}

We present a neural IK model that can capture the inherent one-to-many ambiguity of the problem, while training on a dataset with one-to-one samples. The architecture consists of a single hypernetwork and a sequence of primary matrices. Making use of the hierarchical nature of the problem, each joint is sampled from a GMM that is conditioned on the samples performed for the previous joints in the kinematic chain.

Having an accurate feed forward model that supports multiple outputs has a few advantages, which we demonstrate. First, the solution is extremely efficient at run-time. Second, it entails an effective representation of the Cartesian positions. Third, one can use such a model to obtain multiple smooth paths. In addition, while not demonstrated here, having a differentiable model allows it to be incorporated as a module in a complex network during inference or as part of a loss at training time.

\bibliography{ik}
\bibliographystyle{icml2022}

\end{document}